\documentclass[letterpaper, 10 pt, conference]{ieeeconf}  

\IEEEoverridecommandlockouts                              

\usepackage{multirow}
\usepackage{graphicx}
\usepackage{array}
\usepackage{booktabs}
\usepackage{xcolor}
\usepackage{xspace}
\usepackage{overpic}
\usepackage{makecell}
\usepackage{cite}
\usepackage{comment}

\usepackage{amsmath}      
\usepackage{amssymb}      
\usepackage{amsfonts}     
\usepackage{bm}           
\usepackage{mathtools}    

\usepackage{algorithm}    
\usepackage[noend]{algpseudocode}
\usepackage{listings}     

\usepackage{array}        
\usepackage{booktabs}     

\usepackage{graphicx}     
\usepackage{subcaption}   

\usepackage{hyperref}     

\usepackage{cleveref}     
\usepackage{tcolorbox} 
\usepackage{mathrsfs}
\newtcbox{\mybox}{colframe=black!75!white, colback=white, boxrule=0.8pt, arc=3mm}

\let\labelindent\relax
\usepackage{enumitem}

\def\bl#1{\textcolor{blue}{#1}}
\def\ours{\textsc{KARL}\xspace}

\title{\LARGE \bf
KARL: Kalman-Filter Assisted Reinforcement Learner for Dynamic Object Tracking and Grasping
}

\author{Kowndinya Boyalakuntla\qquad Abdeslam Boularias\qquad Jingjin Yu
\thanks{The authors are with the Department of Computer Science, Rutgers
University, 08854 New Brunswick, USA.}
}

\begin{document}

\maketitle
\thispagestyle{empty}
\pagestyle{empty}

\begin{abstract}
We present \bl{K}alman-filter \bl{A}ssisted \bl{R}einforcement \bl{L}earner (\bl{\ours}) for dynamic object tracking and grasping over eye-on-hand (EoH) systems, significantly expanding such systems' capabilities in challenging, realistic environments. 
In comparison to the previous state-of-the-art, \ours (1) incorporates a novel six-stage RL curriculum that doubles the system’s motion range, thereby greatly enhancing the system's grasping performance, (2) integrates a robust Kalman filter layer between the perception and reinforcement learning (RL) control modules, enabling the system to maintain an uncertain but continuous 6D pose estimate even when the target object temporarily exits the camera’s field-of-view or undergoes rapid, unpredictable motion, and (3) introduces mechanisms to allow retries to gracefully recover from unavoidable policy execution failures. 
Extensive evaluations conducted in both simulation and real-world experiments qualitatively and quantitatively corroborate KARL's advantage over earlier systems, achieving higher grasp success rates and faster robot execution speed.
%
Source code and supplementary materials for \ours will be made available at: {\color{blue}{\url{https://github.com/arc-l/karl}}}.
\end{abstract}

\section{Introduction}\label{sec:intro}

Humans, and animals in general, interact with the physical world through observing and handling everyday objects~\cite{mason2018toward}, which makes object tracking and manipulation arguably the \emph{most fundamental skill} for physical intelligence. 
In robotics, autonomous grasping in \emph{stationary} settings has been extensively studied \cite{kleeberger2020survey,PanZenLiYuHau22TRO}, typically using decoupled vision and manipulation sub-systems where the camera does not move with the manipulator. While effective for static tasks, this approach struggles in dynamic scenarios where objects move or become occluded. Real-world interactions, such as handovers, require continuous tracking and adaptive grasping, highlighting the need for more integrated solutions.

\begin{figure}[t!]
\begin{center}
\begin{overpic}[width=\columnwidth,tics=5]
{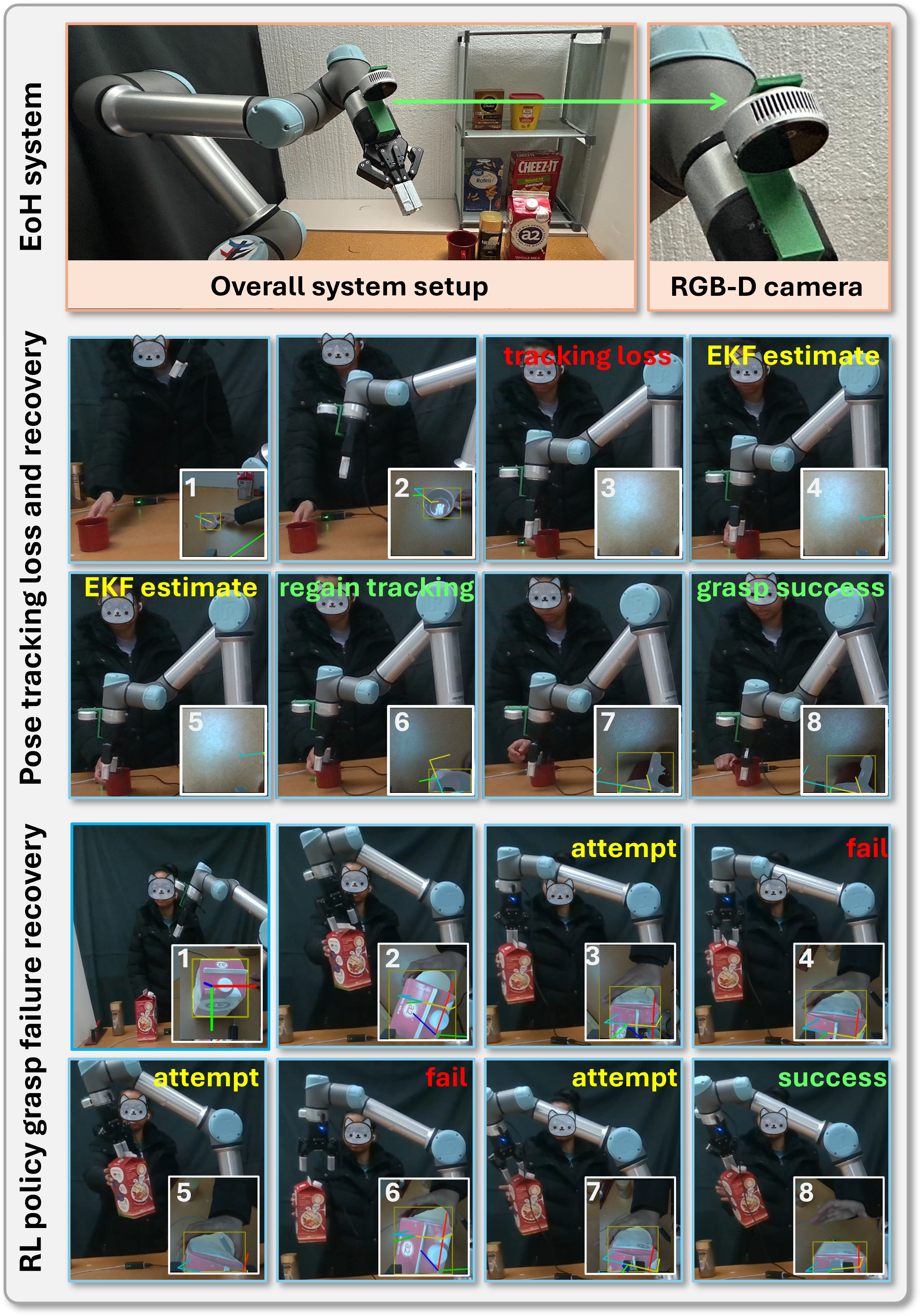}
\end{overpic}
\end{center}
\vspace{-2mm}
\caption{\label{fig:intro} [top row] The EoH system setup used for validating \ours's performance. [middle row] When tracking loss occurs due to an object temporarily going out-of-view or getting occluded, \ours maintains a EKF estimate of the object's pose to allow pose tracking recovery. [bottom row] \ours can detect and gracefully handle grasping failures and make multiple retries until success.}
\vspace{-4mm}
\end{figure}

Through coupling vision and manipulation sub-systems, eye-on-hand (EoH) and eye-in-hand (EiH)\footnote{The difference between EoH and EiH systems are generally minor; in this work, we use eye-on-hand (EoH) to refer to both.} systems \cite{yan2017sim,cheng2022design, vasile2022grasp,earl} can have both the camera and robot's hand follow the target object, avoiding the above-mentioned pitfalls of decoupled systems. EoH systems hold the promise of revolutionizing robotic manipulation across a diverse array of real-world applications. For example, in cluttered and unstructured environments \cite{cheng2022design}, EoH systems make it possible to autonomously track and grasp moving objects with precision, transforming operations in industrial automation and advanced service robotics. EoH systems' ability to provide real‐time visual feedback also makes them particularly suitable for teleoperation, shared autonomy \cite{yan2017sim}, and prosthetic hand control \cite{vasile2022grasp}. Recently, \emph{imitation learning} (IL) systems have largely converged to dual-arm setups where each arm's end-effector has a fixed camera sensor, e.g.,~\cite{fu2024mobile,black2024pi_0}.

Fully leveraging EoH systems presents significant algorithmic and computational challenges, as rapid sensor feedback must be continuously processed for real-time control. 
While end-to-end IL systems \cite{fu2024mobile,black2024pi_0} perform well in stationary scenes, they require extensive data collection and computing power and struggle with dynamic object motion.
Recently, Huang et al. developed the EARL~\cite{earl} reinforcement learning (RL) framework specifically for EoH systems. EARL can track/follow moving objects in real-time and find the right moment to grasp them. EARL achieves this by tightly integrating a perception sub-routine for 6D pose tracking and an RL sub-routine for robot/end-effector control. The pipeline allows EARL to handle a variety of challenging object motion patterns. 

Whereas EARL shows promise in tackling dynamic manipulation settings, due to the need to handle conflicting constraints of EoH systems, it also comes with severe limitations (to be detailed shortly) that prevent it from being practical. To better leverage the capability of EoH systems and render them more practical, in this work, we propose a \emph{Kalman-filter Assisted Reinforcement Learner} (\ours) framework that significantly expands EARL's capabilities. In particular, \ours brings the following key contributions: 
\begin{itemize}[leftmargin=4mm]
\item \ours introduces a Kalman filter layer that sandwiches between the perception and the RL control modules, allowing it to gracefully handle the case where rapid object movement causes the target object to disappear from the camera's field-of-view by maintaining an uncertain object pose estimate. 
EARL completely fails in such cases.
\item EoH systems' strong coupling of camera and robot end-effector poses makes training RL policies that fully utilize the robot arm's motion ranges challenging. This forces EARL to have a limited workspace. \ours addresses this issue by introducing a sophisticated 6-step RL curriculum design to gradually expand the EoH system's reachability. Simultaneously, the majorly updated training schedule boosts the EoH system's speed by over $20\%$.
\item Also by making changes to how the RL policy is trained and executed, \ours possesses the capability to recover from grasp failures and retry multiple times, as long as the target object can be continuously tracked. In contrast, EARL can only make a single grasp attempt. 
\end{itemize}
%
In addition to the above-mentioned major upgrades, we also made cross-the-board updates to the original EARL's (perception and control) components. Notably, we replaced the object tracking module from BundleTrack~\cite{bundletrack} to FoundationPose~\cite{foundationpose} where applicable, leading to a few percentage points of success rate increase over the EARL baseline.


\section{Related Work}\label{sec:related}
\textbf{Approaches to Robotic Grasping}: 
Robotic grasping methods fall into two categories: analytic/geometric and data-driven. Analytic approaches rely on known object models or shape primitives, using geometry or physics-based analyses to plan stable grasps, often leveraging CAD models \cite{du2021vision,wan2020planning,jain2016grasp}. In contrast, learning-based methods generalize to novel objects by training on large grasp datasets \cite{kleeberger2020survey}. These include deep networks that sample and rank grasp candidates from visual input \cite{liang2019pointnetgpd} and reinforcement learning (RL) policies for grasp synthesis \cite{mohammed2020review,zhang2023cherry}. Advances in algorithms and data, such as grasping datasets and simulations, have significantly improved learning-based grasping success.

\textbf{Eye-on-Hand (EoH)  vs. Static Camera Setups}:
Most vision-based manipulators use fixed cameras \cite{tuscher2021deep,huang2022parallel}, simplifying perception but introducing occlusion and coverage limitations. Multi-camera setups help but add complexity. Eye-on-Hand (EoH) architectures mount the camera on the robot’s wrist or end-effector \cite{yan2017sim,cheng2022design,vasile2022grasp,earl}, ensuring continuous object tracking and eliminating blind spots. While beneficial in dynamic scenarios, the tight coupling introduces planning constraints that can limit system motion.


\textbf{Dynamic Grasping}:
demands real-time adaptation to unpredictable object motion. Early approaches relied on feedback control or fast vision networks, such as Morrison et al.’s Generative Grasp CNN for rapid closed-loop grasping in mildly dynamic scenes \cite{morrison2018closing}. Other methods continuously re-planned trajectories using multi-camera tracking \cite{marturi2019dynamic} or motion prediction \cite{akinola2021dynamic}, which struggled with erratic movements and relied on fixed cameras, limiting real-time pursuit.

\textit{Visual Servoing (VS) based grasping using EoH Camera} is a well-established manipulation framework that uses visual feedback to control a robot's motion and perform closed-loop, real-time grasping~\cite{wu2022survey}. One of the main approaches within VS is Position-Based Visual Servoing (PBVS), which estimates the full 6D pose of an object and directly computes the control commands in task space to drive the robot to the desired configuration. PBVS has been used for dynamic grasping of moving objects~\cite{burgess2023architecture} with unpredictable motion, although limited to slow object movement scenarios.
\textit{Deep RL} has emerged as a promising alternative, with Song et al. \cite{song2020grasping} training an RL policy that maps wrist-camera images to Q-values, enabling 6-DoF grasping in clutter. However, it was constrained to discrete actions and slower object motions. Many prior methods imposed constraints on grasp orientation or speed to ensure reliability \cite{kalashnikov2018scalable, song2020grasping}.

\section{Preliminaries}\label{sec:background}

\subsection{Problem Formulation}
We aim to endow EoH systems with the ability to dynamically track and grasp moving target objects. Our primary objective is to enable EoH systems to perform dynamic grasping in full six degrees of freedom (i.e., within the SE(3) space) for a moving object whose motion is a priori unknown. We make minimal assumptions about the target—namely, that it is a graspable rigid body. No assumptions are made regarding the object's shape, identity, or specific motion profile; the object may move freely within the robot's reachable workspace. Successfully solving this problem requires addressing two interrelated challenges:
\begin{itemize}[leftmargin=4mm]
\item Object Tracking and Pursuit: The system must continuously estimate the 6D pose of the moving object and actively control the robot to keep the target within view.
\item Grasp Execution: Simultaneously, the robot must effectively approach and execute a stable grasp on the target object despite its unpredictable motion.
\end{itemize}
A grasp is deemed successful if the system is able to securely seize and lift the target object. 

Besides simulation-based evaluation, we evaluate an EoH system with the Universal UR-5e (a 6-DOF robot) with a mounted Intel RealSense L515 RGB-D camera. A Robotiq 2F-85 two-finger gripper serves as the end-effector.

\subsection{Eye-on-hand Reinforcement Learner (EARL) Overview}
Eye-on-Hand Reinforcement Learner (EARL) \cite{earl} integrates perception and RL-based tracking and grasping for EoH systems, which eliminates workspace constraints and occlusion issues typical to external camera setups. 
In an EoH system, an RGB-D sensor mounted on the robotic arm can track objects in real time. In EARL, to register the 6D pose of the object, a few 3D points are selected manually on the point cloud of the scene, which is then fed to the tracking module that uses R2D2~\cite{r2d2} and BundleTrack~\cite{bundletrack} to maintain keyframe memory and performs 6D pose estimation updates in real-time.
In EARL, a target object is manually selected by drawing a bounding box. It leverages Contact-GraspNet~\cite{contact_grasp_net} to generate multiple 6-DoF grasp candidates. These candidates are stored in a grasp memory pool, allowing EARL to select the most stable and reachable grasp based on the current pose of the object.

EARL trains a Proximal Policy Optimization (PPO)~\cite{ppo} RL policy to generate joint velocity commands and gripper actions, given object pose input from the perception sub-system and the robot's joint states. A three-step curriculum is designed to simultaneously track the object and drive the end-effector's motion. To achieve robust grasping, reward is carefully shaped to: (1) \textbf{maintain object tracking} by assigning a penalty whenever the target object moves out of view, (2) \textbf{avoid collision} through a penalty discouraging pre-grasp collisions while ensuring the gripper remains close, (3) \textbf{encourages proper gripper alignment and approach angles} along multiple axes ($y$,$z$) to maximize grasp stability, and (4) \textbf{execute successful grasps} by rewarding the agent for closing the gripper at the optimal moment.

\textbf{Limitations}.
While EARL's three-stage RL curriculum progressively increases task difficulty, the initial stage imposes overly restrictive starting robot joint positions, restricting object pose diversity. In later stages, strong penalties are imposed to prevent object tracking loss and collision, further reducing the allowed object's motion range and speed. Quantitatively, EARL's effective workspace is confined to a $40 \times  40 \times 40$ $cm^3$ volume and needs a longer time for its RL policy to run (e.g., $> 15$ seconds to reach $90\%$ success rate due to slower supported robot speed. Moreover, EARL lacks mechanisms to recover from RL policy failures, leading to $100\%$ failures if the object quickly moves out of the camera's view or a grasp attempt fails. 
To fully unlock the potential of EoH systems in dynamic manipulation settings, these severe limitations must be overcome.

\section{Proposed System}\label{sec:system}
\begin{figure*}
    \centering
    \begin{overpic}[width=1\linewidth]{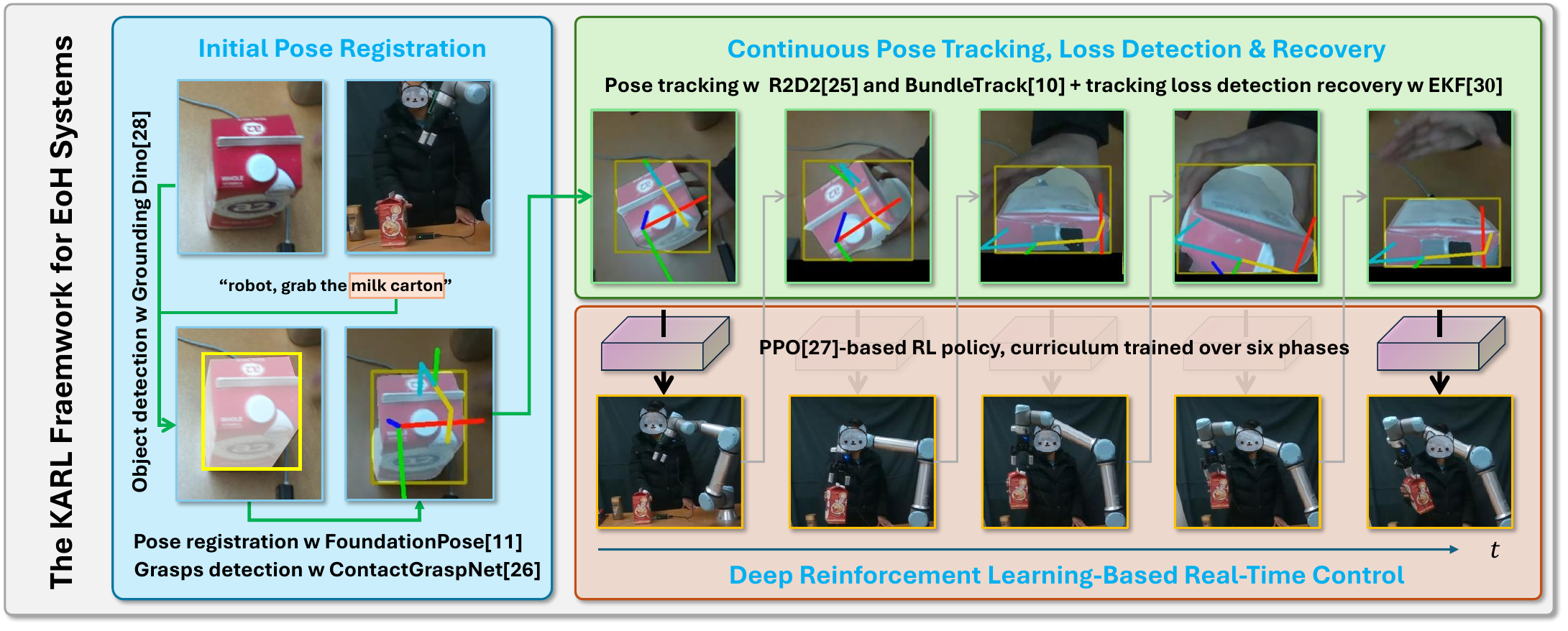}
        \small
    \end{overpic}
    \caption{The \ours framework for EoH systems. It has three interconnected modules for initial pose estimation, continuous pose tracking, and RL-based continuous robot control. In the beginning, based on user input \ours automatically detects the target object, estimates its initial pose, and computes a set of candidate grasps. It then enters the main pose tracking and control loop. In the process, \ours also keeps monitoring and recovering from potential tracking loss (due to sudden quick object movements or temporary occlusion) and possible grasping failure.}
    \label{fig:karl}
    \vspace{-5mm}
\end{figure*}
\ours has three main, interconnected modules (see Fig.~\ref{fig:karl} for an illustration). The perception sub-system of \ours contains two: an initial pose registration module and a continuous pose tracking/recovery module. The third module is the RL-based robot controller. 

\begin{algorithm}
\caption{Active Pose Estimation and Object Grasping}
\label{algo:karl}
\footnotesize
\begin{algorithmic}[1]
\State \textbf{Initialization:}
\State $t \gets 0$, $\mathcal{I}_t, \mathcal{D}_t$ : RGB, Depth images, $\mathcal{C}_t$ : Camera pose 
\State $\mathcal{G}_t^W, \mathcal{G}_t^C$ : Gripper pose in world \& camera frames
\State $\mathcal{O}_\text{pool}^W, \mathcal{O}_\text{pool}^C$ : Grasp pose pool in world \& camera frames
\State $\mathcal{O}_t^C$ : Grasp pose in camera frame
\State $\mathcal{T}_t^W, \mathcal{T}_t^C$ : Object pose in world \& camera frames
\State Read initial RGB-D image: $(\mathcal{I}_0, \mathcal{D}_0)$
\State Read object name $\mathcal{L}$ (e.g., “coffee cup”)
\State $\mathcal{B}_0 \gets \operatorname{GroundingDINO}(\mathcal{I}_0, \mathcal{D}_0, \mathcal{L})$
\State $\mathcal{M}_0 \gets \operatorname{2DTracker_{init}}(\mathcal{I}_0, \mathcal{B}_0)$
\State $\mathcal{K} \gets \operatorname{CAD\_Model}(L)$
\State $\mathcal{T}_0^{C} \gets \operatorname{FoundationPose}(\mathcal{I}_0, \mathcal{D}_0, \mathcal{M}_0, \mathcal{K})$
\State $\mathcal{O}_\text{pool}^{W} \gets \operatorname{ContactGraspNet}(\mathcal{I}_0, \mathcal{D}_0)$
\State $\mathcal{O}_\text{pool}^{C} \gets \operatorname{TransformToCameraFrame}(\mathcal{C}_0, \mathcal{O}_\text{pool}^{W})$
\State $\mathcal{O}_{0}^{C} \gets \operatorname{SelectBestGrasp}(\mathcal{G}_0^C, \mathcal{O}_\text{pool}^C)$
\State $\mathcal{T}_\text{obj\_grasp}^C \gets \operatorname{ComputeTransform}(\mathcal{T}_0^C, \mathcal{O}_0^C)$
\State Initialize EKF with state $\mathbf{x}_0$, covariance matrix $\mathbf{P}_0$
\State $\text{use\_foundationpose} \gets \textbf{False}$
\State $\text{handover} \gets \textbf{False}$

\While {$\text{handover} = \textbf{False}$ and $t < t_{\max}$}
    \State $t \gets t+1$
    \State Read new RGB-D frame: $(\mathcal{I}_t, \mathcal{D}_t)$
    \State $\mathcal{M}_t, \mathcal{B}_t \gets \operatorname{2DTracker_{track}}(\mathcal{I}_t)$

    \If {$\mathcal{M}_t \neq \textbf{None}$}  
        \If {$\text{use\_foundationpose} = \textbf{True}$}  
            \State $\mathcal{T}_t^C \gets \operatorname{FoundationPose}(\mathcal{I}_t, \mathcal{D}_t, \mathcal{M}_t, \mathcal{K}, \mathcal{T}_{t-1}^C)$
                \State $\text{use\_foundationpose} = \textbf{False}$
        \Else  
            \State $\mathcal{T}_t^C \gets \operatorname{BundleTrack\_R2D2}(\mathcal{I}_t, \mathcal{D}_t, \mathcal{M}_t, \mathcal{T}_{t-1}^C)$
        \EndIf
        \State $\mathcal{T}_t^W \gets \operatorname{}(\mathcal{C}_t *\mathcal{T}_t^C)$
        \State $\operatorname{EKF.update}(\mathcal{T}_t^W)$
    \Else  
        \State $\text{use\_foundationpose} \gets \textbf{True}$
        \State $\mathcal{T}_t^W \gets \operatorname{EKF.predict}()$
        \State $\mathcal{T}_t^C \gets \operatorname{}(inverse(\mathcal{C}_t) * \mathcal{T}_t^W)$
    \EndIf  

    \State $\mathcal{O}_t^C \gets \mathcal{T}_\text{obj\_grasp}^C \cdot \mathcal{T}_t^C$
    \State $\mathcal{O}_t^C \gets \operatorname{SelectBestGrasp}(\mathcal{G}_t^C, \mathcal{O}_t^C)$
    \State Move robot gripper toward $\mathcal{O}_t^C$

    \If {$\operatorname{Align}(\mathcal{G}_t^C, \mathcal{O}_t^C) = \textbf{True}$ and $\operatorname{Dist}(\mathcal{G}_t^C, \mathcal{O}_t^C) < \epsilon$}
        \State Grasp the object and lift
        \If {Grasping Failed} 
            \State Invoke RL Policy and continue
        \Else
            \State Grasping Success!
            \State $\text{handover} \gets \textbf{True}$
        \EndIf
    \EndIf  
\EndWhile
\end{algorithmic}

\end{algorithm}
Algorithm~\ref{algo:karl} outlines the method.

\subsection{Initial Pose Registration}
\ours takes in requests in the form of simple natural language inputs, e.g., ``robot, grab that \emph{milk carton}.'' Upon receiving a request, \ours prompts Grounding DINO~\cite{gdino} to automatically extract the target object from images taken from the hand-mounted RGB-D camera. At this point ($t = 0$), two additional tasks are performed: initial pose estimation/registration and candidate grasp generation. For the former, FoundationPose~\cite{foundationpose} is invoked to perform the pose estimation of the target object using RGB-D input, object mask and its textured mesh. If the CAD model of the object is not available, we utilize BundleSDF~\cite{wen2023bundlesdf} by feeding a sequence of reference RGB-D images of the target and mask of the first frame in the sequence. 
For the latter, Contact-GraspNet~\cite{sundermeyer2021contact} is applied directly to the (segmented) depth data to produce (a few) candidate grasps. Formally, given RGB-D, mask and CAD model as input $(\mathcal{I}_0, \mathcal{D}_0, \mathcal{M}_0, \textcolor{blue}{\mathcal{K}})$, the initial pose registration phase yields object's 6D pose $\mathcal{T}_0^{\mathcal{C}}$, and grasp pose candidates $\mathcal{O}_\text{pool}^{\mathcal{W}}$ in the camera $\mathcal{C}$ and world  $\mathcal{W}$ frames of reference respectively.
\begin{figure*}
    \centering
    \begin{overpic}[width=1\linewidth]{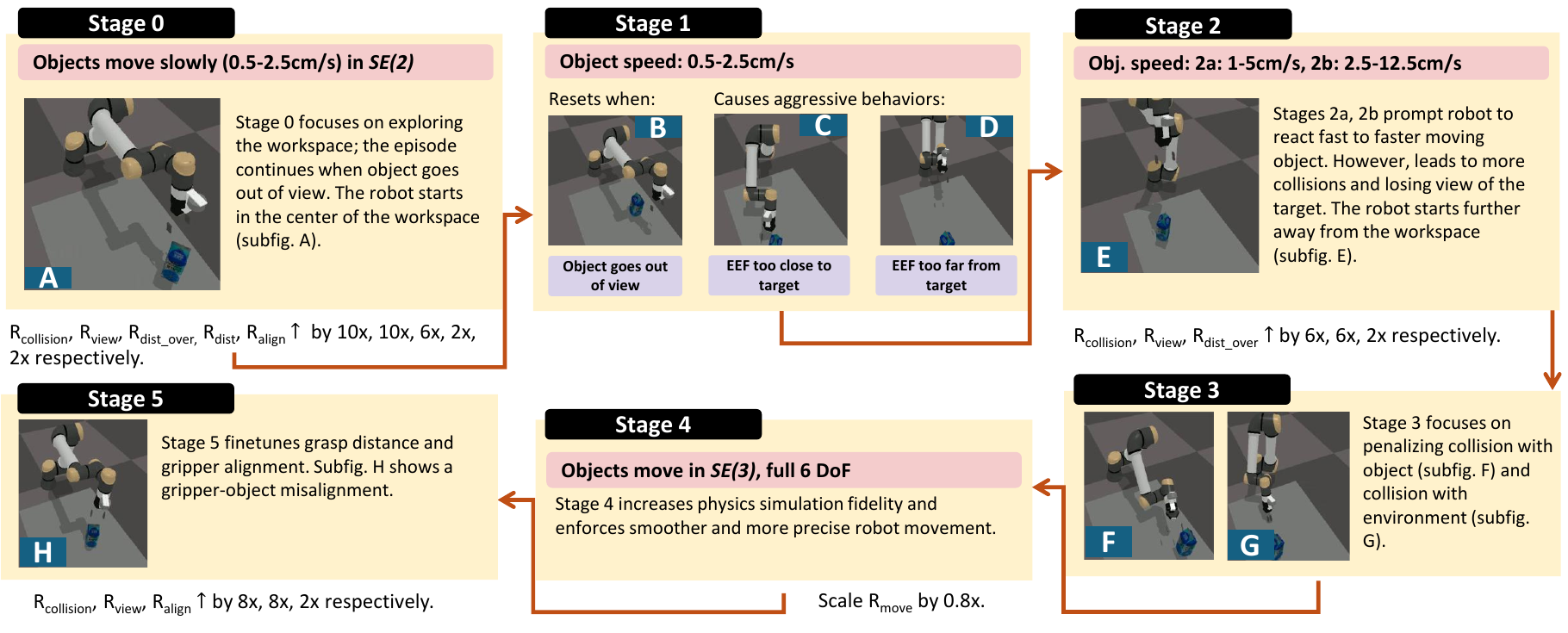}
        \small
    \end{overpic}
    \caption{The six stage curriculum used for training \ours's MLP network for RL-based robot control, using PPO~\cite{ppo}.}
    \label{fig:curriculum}
\end{figure*}
\subsection{Real-Time Pose Tracking and Tracking Loss Recovery}
\subsubsection{Continuous Pose Tracking Before Tracking Loss} 
Starting from initial pose $\mathcal{T}_0^{\mathcal{C}}$ given by FoundationPose~\cite{foundationpose}, we let the pose tracker (R2D2~\cite{r2d2} + BundleTrack~\cite{bundletrack} from~\cite{earl}) track the object's pose $\mathcal{T}_t^{\mathcal{C}}$ in real-time. Ideally, FoundationPose can be used to also track the object, however in our EoH setup both the camera and target object move randomly in 3D space. As a result, we found that the tracker in FoundationPose often becomes unstable. We attribute this to its pose refiner model encountering object trajectories (from our setup) that were likely absent from its training data. 
Finetuning FoundationPose for object trajectories in our setup can be explored in the future to streamline the design. 

\subsubsection{Tracking Loss Detection and Recovery} 
During the pose tracking process, an Extended Kalman Filter (EKF)~\cite{ekf} is maintained to provide a secondary pose estimate ($\mathcal{T}_{t}^{\mathcal{W}}$, $\mathcal{W}$ denotes the world coordinate frame.). We use a relatively standard EKF implementation to track the target pose, essentially using $x_{t+1} = x_t + v_t\Delta t$ for the process update (here, $x_k = \mathcal T_t^{\mathcal{W}}$ and $v_t$ are the proper linear and rotational velocities computed using known pose history $\mathcal{T}_{0:t-1}^{\mathcal{W}}$). 
When an object mask can be reliably detected, this EKF estimate is not used for downstream tasks. 
When at a time $t$, the target object leaves the FOV of the camera or gets occluded, we lose the mask ($\mathcal{M}_t$) of the object and this signals \ours to switch to using EKF estimated pose. 

During the period where the tracked pose is provided by the EKF, because no new pose observation is available, the last known observation is used to update the EKF, which makes sense for both the case when the object moves out of view and when it is occluded by another object. This pushes the EKF's estimate to grow more and more uncertain over time, which is expected. The growing uncertainty naturally drives the control policy (to be detailed in Sec.~\ref{sec:controller}) to behave more conservatively, which is the desired behavior. In particular, we sample from the uncertain pose estimates a pose that maximizes the expected view of the uncertain regions so that the camera can get the object back in view. We note that, as with any recovery method, the duration of the tracking loss cannot be very large. Otherwise, the RL policy can fail. 

BundleTrack works when the pose changes between two frames are relatively small. When the object comes back to view (at some time $t'$), BundleTrack generally cannot recover the pose of the object. When this happens, the pose of the object must be \emph{re-initialized} for tracking. As the mask $\mathcal{M}_{t'}$ of the object is recovered by the 2DTracker~\cite{2dTracker}, we invoke FoundationPose again to quickly re-register the pose only once. The new pose is denoted as $\mathcal{T}_{t'}^{\mathcal{C}}$. We convert this pose to the world coordinate frame as $\mathcal{T}_{t'}^{\mathcal{W}}$, and continuously recompute the object pose in camera coordinates as the robot continues to reach and grasp. Since the robot is moving and hence the camera is also moving, pose estimation needs to be both quick and accurate. We feed FoundationPose, the current mean of pose distribution from the EKF ($\mathcal{T}_{EKF(t, \mu)}^{\mathcal{C}})$, after converting to camera coordinates), and reduce the number of object 6D pose hypotheses to under 60. This estimation is both fast and accurate. 

\subsection{RL-Based Robot Controller}\label{sec:controller}
Similar to \cite{earl}, we train the RL agent, a multi-layer perception (MLP) with PPO\cite{ppo},
\[
[\mathcal{Q}_v^{t+1}, b^{t+1}] = MLP(\mathcal{P}_g^{t}, \mathcal{E}_g^{t}, \Delta_g^{t}, \mathcal Q_p^{t}, \mathcal Q_v^{t}, b^{t}),
\]
where $Q_v$ is the joint velocities, $b$ is the gripper open/close action, $\mathcal{P}_g$ are gripper keypoints, $\mathcal{E}_g = \mathcal{P}_g - \mathcal{P}_o$ is the positional error between the gripper and object grasp keypoints (to be minimized for a successful grasp), $\Delta_g = \mathcal{P}_o - \mathcal{P}_o'$ reflects the change in object grasp keypoints between consecutive frames, and $\mathcal{Q}_p$ denotes joint positions, normalized based on joint limits.
%

\subsubsection{Reward Shaping and Curriculum Design}
To accommodate the EoH system's intrinsic constraints and support the desired dynamics behavior, a reward structure and curriculum of similar complexity are required. \ours settled on an overall reward of the following form: 
\begin{equation}
\begin{split}
\mathcal{R}^{(i)} = 
\mathcal{R}_{grasp} + 
\lambda_{dist}^{(i)} \mathcal{R}_{dist} + 
\lambda_{dist\_over}^{(i)} \mathcal{R}_{dist\_over}  \\
+ \lambda_{align}^{(i)} \mathcal{R}_{align} + 
 \lambda_{collision}^{(i)} \mathcal{R}_{collision} + 
\lambda_{view}^{(i)} \mathcal{R}_{view}  \\
+ \lambda_{gripper}^{(i)} \mathcal{R}_{gripper\_penalty} + 
\lambda_{move}^{(i)} \mathcal{R}_{move},
\end{split}
\end{equation}
%
where terms denote \emph{successful grasping reward}, \emph{distance reduction reward}, \emph{not staying overly distant reward}, \emph{gripper alignment reward}, \emph{collision avoidance reward}, \emph{object visibility reward}, \emph{premature gripper closure avoidance reward}, and \emph{stable gripping reward}, in that order. For more details and rationale of the reward terms, see~\cite{earl}.
At stage $i$ in the curriculum, the weight coefficients $\lambda^{(i)}$ are adjusted to guide the learning process progressively.
The reward structure discourages inefficiencies while reinforcing effective grasping behavior.  


\textbf{Curriculum Design}
We settle down on a six-stage curriculum (see Fig.~\ref{fig:curriculum}) with smoother performance-adaptive transitions between stages and improved reward shaping, with earlier stages being more conservative. In stage 0, the robot starts in the workspace center with closely sampled target object poses. The target object moves only in $SE(2)$ until stage 3. Stage 0 lets the robot freely explore without stopping it, even if the object goes out of view. In stage 1, object visibility is enforced more aggressively (losing visibility triggers environment reset), ensuring the robot learns to maintain sight of the object. Instead of gradually increasing object speeds in training, we accelerate the object movements earlier in \ours, to promote faster adaptation to higher curriculum levels with faster object movements. The system also increases rewards for firm grasps upon correct alignment. In summary, for early stages (0-1), high penalties are enforced for losing object visibility and collisions and moderate rewards for grasping. Mid-stages (2-4) focus on balancing speed and safety, with increasing importance on precise grasping (in stage 4, the robot operates with high control frequency to ensure precise robot movements). In the later stages, alignment and action penalties have been fine-tuned to discourage unnecessary movements when the robot is about to close the gripper near the target, improving efficiency and reducing grasp failures due to misalignment. The curriculum was trained in parallel over 8192 Isaac Gym~\cite{isaacgym} environments on a RTX 4090 GPU. 


\subsubsection{Grasp Failure Detection and Recovery}
During real-world experimentation, the object may slip from the gripper due to factors such as inadequate grip force, object inertia, or external disturbances. To address this issue, we introduced a grasp failure detection and recovery mechanism within the control loop of the real robot policy.

Instead of assuming a successful grasp upon closing the gripper, the system continuously monitors the gripper's position to determine whether the object has been securely held. If the gripper's position remains above a predefined threshold (indicating that the object has not been grasped), the robot immediately reopens the gripper and attempts to reposition itself for another grasp.

We modified the control loop to:
(1) continuously provide new pose observations by re-evaluating the target object's position and orientation in real time,
(2) invoke the RL policy iteratively to adapt the robot's motion and attempt a re-grasp until success,
and (3), introduce a brief stabilization delay before lifting the object to prevent premature slippage. 
Overall, this ensures more reliable object manipulation in dynamic environments.

\section{Evaluation}\label{sec:evaluation}
To comprehensively evaluate our method, we conducted six experiments—five in simulation (\textbf{A-E}) and one on a physical robot (\textbf{F}), with varying complexity through:
\begin{itemize}[leftmargin=4mm]
    \item \textbf{A}: Faster object speeds.
    \item \textbf{B}: Stricter grasping task time constraints.
    \item \textbf{C}: Expanded target workspace.
    \item \textbf{D}: Tracking loss scenarios.
    \item \textbf{E}: Combined effect of A–D in an ablation study.
    \item \textbf{F}: Real-world evaluation using a UR5e robotic arm.
\end{itemize}
\subsection{Target Object Motion}
The target object, initialized at a workspace boundary, may follow the following motions:
\begin{itemize}[leftmargin=4mm]
    \item \textbf{Linear (Regular)}: Constant speed (0–5 cm/s), no rotation. Used in $70\%$ of trials in Experiments B, C.
    \item \textbf{Linear (Fast)}: Higher speed (up to 15 cm/s), zero rotational velocity.
    \item \textbf{Random}: Translational velocity (0–5 cm/s) with random rotations ($0^\circ$–$14.5^\circ$) along X, Y, and Z axes. Used in $30\%$ of trials in Experiments B and C.
    \item \textbf{Disruptive}: Sudden velocity spikes ($>30$ cm/s), causing temporary tracking loss.
\end{itemize}

\subsection{Higher Object Speeds}
For increasing linear motion speeds (3 cm/s to 15 cm/s), \ours maintained a near-perfect success rate ($>$99\%) with zero collisions for speeds up to 9 cm/s. Overall, \ours outperformed EARL, achieving a 92.58\% success rate. While EARL struggled with tracking failures at higher speeds, KARL's EKF module prevented these issues, ensuring robust object tracking and grasp execution. (See Table~\ref{tab:obj_speed}).
 
\setlength{\tabcolsep}{4.1pt}
\begin{table}[htbp]
\caption{Performance comparison of EARL and KARL under different target object speeds (1000 Isaac Gym Envs)}
\renewcommand{\arraystretch}{0.85}
\begin{tabular}{llccccccc}
\toprule
\multirow{2}{*}{\;} & \multirow{2}{*}{Policy} & \multicolumn{6}{c}{Target Object Speed (cm/s)} \\ 
\cmidrule(lr){3-8} 
                        &                         & 3  & 6  & 9  & 12  & 15   & Random \\ 
\midrule
\multirow{2}{*}{Success ($\uparrow$)} & EARL                     & 94.38   & 95.40   &  95.83  &  80.22  &  64.64  &  89.21  \\  
                         & KARL              & \textbf{99.21}   & \textbf{99.33}   &  \textbf{99.10}  & \textbf{82.81}   & \textbf{67.34}   & \textbf{92.58}   \\ 
\midrule
Time Limit  & EARL                 & \textbf{0.45}   &  1.35  & \textbf{1.13}   & \textbf{1.01}   & \textbf{0.68}   & \textbf{0.79}   \\  
(35 sec) ($\downarrow$)                                     & KARL       & 0.79   &  \textbf{0.67}  & 0.90   &  8.65  & 14.08   &  3.82  \\  
\midrule
\multirow{2}{*}{Collision ($\downarrow$)} & EARL                   &  0.11  & 0.22  & 0.11   & \textbf{3.93}   & \textbf{5.63}   & \textbf{1.12}   \\  
                           & KARL         & \textbf{0.00}   &  \textbf{0.00}  & \textbf{0.00}   & 8.54   &  18.58  & 3.60   \\  
\midrule
Tracking  & EARL            &  5.06  & 3.03   & 2.93   & 14.83   &  29.05  &  8.88  \\  
Failure ($\downarrow$)          & KARL     & \textbf{0.00}   & \textbf{0.00}   & \textbf{0.00}   & \textbf{0.00}   & \textbf{0.00}   & \textbf{0.00}   \\  
\midrule
Total & EARL              &  5.62  & 4.60   & 4.17   & 19.78   & 35.36   &  10.79  \\  
Failure ($\downarrow$)                               & KARL       & \textbf{0.79}   &  \textbf{0.67}  & \textbf{0.90}   & \textbf{17.19}   & \textbf{32.66}   &  \textbf{7.42}  \\  
\bottomrule
\end{tabular}
\label{tab:obj_speed}
\end{table}

\begin{table}[htbp]
\caption{Performance comparison of EARL and KARL under different time limits (1000 Isaac Gym Envs)}
\centering
\renewcommand{\arraystretch}{0.85}
\begin{tabular}{lcc}
\toprule
\textbf{Metric (\%)\;} & \textbf{EARL (\%)} & \textbf{KARL (Ours) (\%)} \\  
\midrule
Success ($t \leq 35s$) ($\uparrow$) & 95.16 & \textbf{96.89} \\
Success ($t \leq 30s$) ($\uparrow$) & 93.61 & \textbf{96.34} \\
Success ($t \leq 25s$) ($\uparrow$) & 93.54 & \textbf{96.17} \\
Success ($t \leq 20s$) ($\uparrow$) & 91.91 & \textbf{94.14} \\
Success ($t \leq 15s$) ($\uparrow$) & 84.46 & \textbf{91.83} \\
Success ($t \leq 10s$) ($\uparrow$) & 21.03 & \textbf{26.70} \\
Success ($t \leq 5s$) ($\uparrow$) & \textbf{0.00} & \textbf{0.00} \\
\midrule
Timeout ($>35s$) ($\downarrow$) & 2.70 & \textbf{2.32} \\
Collision ($\downarrow$) & \textbf{0.79} & \textbf{0.79} \\
Tracking Failure ($\downarrow$) & 1.35 & \textbf{0.00} \\
Total Failure ($\downarrow$) & 4.84 & \textbf{3.11} \\
\bottomrule
\end{tabular}
\vspace{-7mm}
\label{tab:time_limit}
\end{table}
\subsection{Decreased Task Time limits}
We evaluated the grasping success of KARL under progressively tighter time constraints, ranging from 35 to 5 seconds (see TABLE~\ref{tab:time_limit}). KARL achieves $91.83\%$ success in 15 seconds, outperforming EARL, which reaches a similar performance in 20 seconds, a 5-second advantage. Notably, $99.9\%$ of KARL’s successful grasps occurred within 25 seconds, consistently surpassing EARL. Interestingly, neither KARL nor EARL succeeded in completing the grasping task within 5 seconds. However, beyond this extreme constraint, KARL also exhibited greater robustness in time-constrained grasping scenarios.

\setlength{\tabcolsep}{5.7pt}
\begin{table}[htbp]
\caption{Performance comparison of EARL and KARL on the enlarged workspace $W_{KARL} = W_{EARL} + W_A + W_B$ (1000 Isaac Gym Envs)}
\centering
\renewcommand{\arraystretch}{0.85}
\begin{tabular}{llcccccc}
\toprule
\multirow{2}{*}{Metric (\%)} & \multirow{2}{*}{Policy} & \multicolumn{5}{c}{Target Object Workspace} \\ 
\cmidrule(lr){3-7} 
                        &                         & $W_A$  & $W_B$  & $W_{EARL}$  &  $W_{KARL}$ \\ 
\midrule
\multirow{2}{*}{Success ($\uparrow$)} & EARL                   &  78.01  & 84.24   & 95.09   & 81.67   \\  
                         & KARL                 &  \textbf{97.16}  &  \textbf{97.41}   & \textbf{96.89}   & \textbf{90.47}   \\  
\midrule
Timeout & EARL          & 11.45   & 3.28  &  2.77  & 12.04     \\  
(35 sec) ($\downarrow$)                                     & KARL      & \textbf{2.03}   &  \textbf{1.85}  & \textbf{2.32}   & \textbf{7.10}   \\  
\midrule
\multirow{2}{*}{Collision ($\downarrow$)} & EARL                  & 1.53   &  1.67  & \textbf{0.79}   & \textbf{2.27} &   \\  
                           & KARL              &  \textbf{0.81}  &  \textbf{0.74}  & \textbf{0.79}   & 2.43   \\  
\midrule
Tracking  & EARL           & 9.01   & 10.81   & 1.35   & 4.02  \\  
Failure ($\downarrow$)                                  & KARL          & \textbf{0.00}    &  \textbf{0.00}   &  \textbf{0.00}   & \textbf{0.00}   \\  
\midrule
Total  & EARL              &  21.99  &  15.76  & 4.91  & 18.33   \\  
Failure ($\downarrow$)                               & KARL             &  \textbf{2.84}  &  \textbf{2.59}  & \textbf{3.11}  & \textbf{9.53}   \\  
\bottomrule
\end{tabular}
\vspace{-3mm}
\label{tab:workspace}
\end{table}

\subsection{Expanded Workspace}
We expanded EARL's workspace and doubled the operational volume by adding two new partitions. 
We expanded EARL's target object workspace $W_{EARL}$ by considering two adjacent workspaces $W_A$ and $W_B$.
\ours's RL curriculum roughly doubles the workspace to (40 x 70 x 40 $cm^3$). EARL struggled, while KARL sustained a $\sim97\%$ success rate across all partitions. In the expanded workspace, where the effective range of horizontal motion is nearly doubled, KARL, as expected, experienced increased timeouts but still outperformed EARL across all metrics. This proves its adaptability in larger, more challenging workspaces (refer Table~\ref{tab:workspace}).

\setlength{\tabcolsep}{4.8pt}
\begin{table}[htbp]
\caption{Performance comparison of EARL and KARL on regular and tracking loss scenarios (1500 Isaac Gym Envs)}
\centering
\renewcommand{\arraystretch}{0.85}
\begin{tabular}{llcccccc}
\toprule
\multirow{2}{*}{Metric (\%)} & \multirow{2}{*}{Policy} & \multicolumn{5}{c}{Scenarios (\#)} \\ 
\cmidrule(lr){3-7} 
                        &                         & \makecell{Line \\ (700)}  & \makecell{Random \\ (300)}  & \makecell{Tracking \\ Loss (500)}  & \makecell{Overall \\ (1500)} \\ 
\midrule
\multirow{2}{*}{Success ($\uparrow$)} & EARL                   &  94.60  & 96.00   & 7.10   & 65.72   \\  
                         & KARL                  &  \textbf{96.10}  &  \textbf{98.50}   & \textbf{94.30}   & \textbf{95.98}   \\  
\midrule
Timeout & EARL          & 3.70   & 0.60  &  \textbf{0.00}  & \textbf{1.85}     \\  
(35 sec) ($\downarrow$)                                     & KARL     & \textbf{3.10}   &  \textbf{0.50}  & 5.20   & 3.28   \\  
\midrule
\multirow{2}{*}{Collision ($\downarrow$)} & EARL                  & 1.50   &  2.40  & 1.80   & 1.17 &   \\  
                           & KARL              &  \textbf{0.80}  &  \textbf{1.00}  & \textbf{0.50}   & \textbf{0.74}   \\  
\midrule
Tracking & EARL           & 1.50   & 1.00   & 91.10   & 31.27  \\  
Failure ($\downarrow$)                                  & KARL          & \textbf{0.00}    &  \textbf{0.00}   &  \textbf{0.00}   & \textbf{0.00}   \\  
\midrule
\multirow{2}{*}{Total Failure ($\downarrow$)} & EARL              &  5.40  &  4.00  & 92.90  & 34.29   \\  
                               & KARL            &  \textbf{3.90}  &  \textbf{1.50}  & \textbf{5.70}  & \textbf{4.02}   \\  
\bottomrule
\end{tabular}
\label{tab:tracking_loss}
\end{table}

\subsection{Tracking Loss and Recovery}
Beyond the workspace expansion and the tighter time constraints, we further assessed KARL’s resilience to objects exiting the camera’s view—one of the unavoidable circumstances in dynamic grasping. 
TABLE~\ref{tab:tracking_loss} shows that while EARL performs well under regular motion (94.60\% success), its success drops to 7.10\% in tracking loss scenarios due to its inability to recover lost object poses. In contrast, KARL, leveraging EKF, accurately predicts object motion, achieving 94.30\% success even under disruptions. 
The robot follows EKF's predicted path to the target. However, EKF's update size cannot be tuned for different object speeds, and the robot might reach the end of the episode before reaching the object, as observed in KARL's timeout rate. Despite this, KARL has zero tracking failures and consistently outperforms EARL across all scenarios.

\setlength{\tabcolsep}{3pt}
\begin{table}[]
\caption{Ablation study (large workspace, high object speeds, tracking loss scenarios) (1000 Isaac Gym Envs)}
\label{tab:ablation}
\centering
\begin{tabular}{lccccc}
\toprule
\multirow{2}{*}{Policy} & \multicolumn{5}{c}{Metric (\%)}                                  \\
                        & Success ($\uparrow$) & Timeout($\downarrow$)& Collision($\downarrow$)& 
                        \makecell{Tracking \\Failure($\downarrow$)}& \makecell{Total\\Failure($\downarrow$)}\\
\midrule
EARL                    & 66.29   & 7.84    & 2.54      & 23.32            & 33.71         \\
Old Cur. & 76.70   & 15.21   & 8.08      & \textbf{0.00}       & 23.30         \\
No EKF            & 72.62   & \textbf{4.00}    & \textbf{2.27}      & 21.11            & 27.38         \\
KARL                     & \textbf{83.46}   & 10.46   & 6.08      & \textbf{0.00}       & \textbf{16.54}         \\
\bottomrule
\end{tabular}
\vspace{-4mm}
\end{table}

\subsection{Ablation Study on KARL}
We now assess KARL on a demanding scenario where all the three scenarios in \textbf{Experiments B, C, D} are combined. 70\% of the time, the object moves in linear motion (fast) and is forced out of view with a 30\% probability (disruptive object motion). We did not choose to further increase the out-of-view probability in order to keep the simulation realistic. For the remaining 30\% of trials, the object follows a random trajectory, where it may exit the camera's view inadvertently.

To assess KARL’s key components, we conducted an ablation study by systematically removing key components—EKF (for tracking loss recovery) and the new curriculum (for better grasp adaptation)—to isolate their contributions.
As shown in Table \ref{tab:ablation},
KARL without the new curriculum (but with EKF) eliminates tracking failures and improves upon EARL by a 10.00\% higher success rate, but timeouts remain a challenge. This is partially due to EKF update size limitations and partially due to the lack of imprecise gripper alignment avoidance in the previous curriculum with the object's grasp pose.
    KARL without EKF (but with the new curriculum) reduces gripper misalignment issues, decreasing timeouts and improving the success rate by 6.00\% over EARL. However, without EKF, tracking failures remain high, only slightly lower than EARL.
    KARL achieves the highest success rate (83.46\%), improving upon EARL by 17.17\%, with zero tracking failures and a significantly lower overall failure rate.
These results reinforce the importance of both EKF for robust object tracking recovery and the new curriculum for improved grasp alignment and success, demonstrating that their combined effect enables KARL to outperform EARL across all difficulty levels.

\begin{table}[htbp]
\caption{Real-world comparison of EARL and KARL}
\label{tab:real_exp}
\centering
\begin{tabular}{lllll}
\toprule
\multirow{2}{*}{Policy} & \multicolumn{2}{c}{Success Rate (\%)} & \multicolumn{2}{c}{Premature Grasping Rate (\%)} \\
\cmidrule(lr){2-3} \cmidrule(lr){4-5}
                        & Regular                   & Complex                  & Regular            & Complex            \\
\midrule
EARL                    & 60.00                       & 53.30                     & -                 & -                  \\
\textbf{KARL}           & \textbf{93.30}            & \textbf{80.00}              & \textbf{0.00}        & \textbf{6.60}       \\
\bottomrule
\end{tabular}
\end{table}

\subsection{Real Robot Experiments}
To validate KARL’s real-world performance, we conducted grasping experiments using a UR5e robotic arm equipped with a 2F-85 Robotiq Gripper. Both KARL and EARL were tested in 30 trials, totaling 60 trials across different scenarios:
\begin{itemize}[leftmargin=4mm]
    \item \textbf{Regular Scenes (15 trials):} Only the target object was present, with no obstacles.
    \item \textbf{Complex Scenes (15 trials):} The target moved behind obstacles such as shelves or walls, requiring the robot to follow and maintain visibility, highlighting the advantage of an Eye-on-Hand (EoH) system. However, explicit collision avoidance was not in place, introducing the potential for unintended contact.
\end{itemize}
The target object's motion followed one of three patterns, with 14 trials with linear (regular) motion, 6 trials with random motion, and 10 trials where tracking was intentionally lost, forcing the robot to recover the object’s location.
As shown in Table \ref{tab:real_exp}, KARL outperforms EARL in both regular and complex scenarios. In obstacle-free scenes, KARL achieves an impressive 93.30\% success rate, compared to 60.00\% for EARL. In complex environments, KARL maintains a strong 80.00\% success rate, whereas EARL drops to 53.30\%. EARL falters mainly due to its inability to recover objects that move out of view.

One observed limitation of KARL is premature grasping—in 6.60\% of complex scene trials, KARL attempted to grasp an estimated pose from its EKF-based predictions before reaching the actual object. Unlike in simulation, where successful grasping is confirmed through object contact, our real-world setup lacks this feature, leading to occasional early grasp attempts.

These results reinforce the findings from the simulation studies—KARL’s EKF-based tracking and adaptive grasping curriculum provide significant advantages, especially in dynamic and occlusion-heavy environments. While minor real-world limitations exist, KARL consistently outperforms EARL across all tested conditions, demonstrating superior robustness and adaptability in real-world grasping tasks.

\section{Conclusion}\label{sec:conclusion}
In this work, we introduced KARL—a Kalman-filter Assisted Reinforcement Learner that significantly enhances eye-on-hand (EoH) systems for dynamic object tracking and grasping. By integrating an Extended Kalman Filter, KARL maintains continuous 6D pose estimates even when the target temporarily exits the camera’s view or moves unpredictably. Coupled with a novel six-stage reinforcement learning curriculum, KARL doubles the operational workspace and improves grasp performance through rapid recovery and multiple retry attempts. Evaluations in simulated and real-world environments demonstrate that KARL outperforms previous methods in grasp success rate, execution speed, and collision avoidance. \ours's combined state estimation and adaptive curriculum allow it to handle a broader range of object speeds and motion patterns, enhancing overall reliability in dynamic scenarios. Future work will focus on further refining responsiveness under extreme motions, rendering \ours an essential tool for versatile robotic manipulation.

\bibliographystyle{IEEEtran}

\begin{thebibliography}{10}
\providecommand{\url}[1]{#1}
\csname url@samestyle\endcsname
\providecommand{\newblock}{\relax}
\providecommand{\bibinfo}[2]{#2}
\providecommand{\BIBentrySTDinterwordspacing}{\spaceskip=0pt\relax}
\providecommand{\BIBentryALTinterwordstretchfactor}{4}
\providecommand{\BIBentryALTinterwordspacing}{\spaceskip=\fontdimen2\font plus
\BIBentryALTinterwordstretchfactor\fontdimen3\font minus \fontdimen4\font\relax}
\providecommand{\BIBforeignlanguage}[2]{{%
\expandafter\ifx\csname l@#1\endcsname\relax
\typeout{** WARNING: IEEEtran.bst: No hyphenation pattern has been}%
\typeout{** loaded for the language `#1'. Using the pattern for}%
\typeout{** the default language instead.}%
\else
\language=\csname l@#1\endcsname
\fi
#2}}
\providecommand{\BIBdecl}{\relax}
\BIBdecl

\bibitem{mason2018toward}
M.~T. Mason, ``Toward robotic manipulation,'' \emph{Annual Review of Control, Robotics, and Autonomous Systems}, vol.~1, no.~1, pp. 1--28, 2018.

\bibitem{kleeberger2020survey}
K.~Kleeberger, R.~Bormann, W.~Kraus, and M.~F. Huber, ``A survey on learning-based robotic grasping,'' \emph{Current Robotics Reports}, vol.~1, no.~4, pp. 239--249, 2020.

\bibitem{PanZenLiYuHau22TRO}
Z.~Pan, A.~Zeng, Y.~Li, J.~Yu, and K.~Hauser, ``Algorithms and systems for manipulating multiple objects,'' \emph{IEEE Transactions on Robotics}, vol.~39, no.~1, pp. 2--20, 2022.

\bibitem{yan2017sim}
M.~Yan, I.~Frosio, S.~Tyree, and J.~Kautz, ``Sim-to-real transfer of accurate grasping with eye-in-hand observations and continuous control,'' \emph{arXiv preprint arXiv:1712.03303}, 2017.

\bibitem{cheng2022design}
L.-W. Cheng, S.-W. Liu, and J.-Y. Chang, ``Design of an eye-in-hand smart gripper for visual and mechanical adaptation in grasping,'' \emph{Applied Sciences}, vol.~12, no.~10, p. 5024, 2022.

\bibitem{vasile2022grasp}
F.~Vasile, E.~Maiettini, G.~Pasquale, A.~Florio, N.~Boccardo, and L.~Natale, ``Grasp pre-shape selection by synthetic training: Eye-in-hand shared control on the hannes prosthesis,'' in \emph{2022 IEEE/RSJ International Conference on Intelligent Robots and Systems (IROS)}.\hskip 1em plus 0.5em minus 0.4em\relax IEEE, 2022, pp. 13\,112--13\,119.

\bibitem{earl}
B.~Huang, J.~Yu, and S.~Jain, ``Earl: Eye-on-hand reinforcement learner for dynamic grasping with active pose estimation,'' in \emph{2023 IEEE/RSJ International Conference on Intelligent Robots and Systems (IROS)}.\hskip 1em plus 0.5em minus 0.4em\relax IEEE, 2023, pp. 2963--2970.

\bibitem{fu2024mobile}
Z.~Fu, T.~Z. Zhao, and C.~Finn, ``Mobile aloha: Learning bimanual mobile manipulation with low-cost whole-body teleoperation,'' \emph{arXiv preprint arXiv:2401.02117}, 2024.

\bibitem{black2024pi_0}
K.~Black, N.~Brown, D.~Driess, A.~Esmail, M.~Equi, C.~Finn, N.~Fusai, L.~Groom, K.~Hausman, B.~Ichter \emph{et~al.}, ``$\pi_0 $: A vision-language-action flow model for general robot control,'' \emph{arXiv preprint arXiv:2410.24164}, 2024.

\bibitem{bundletrack}
B.~Wen and K.~Bekris, ``Bundletrack: 6d pose tracking for novel objects without instance or category-level 3d models,'' in \emph{2021 IEEE/RSJ International Conference on Intelligent Robots and Systems (IROS)}.\hskip 1em plus 0.5em minus 0.4em\relax IEEE, 2021, pp. 8067--8074.

\bibitem{foundationpose}
B.~Wen, W.~Yang, J.~Kautz, and S.~Birchfield, ``Foundationpose: Unified 6d pose estimation and tracking of novel objects,'' in \emph{Proceedings of the IEEE/CVF Conference on Computer Vision and Pattern Recognition}, 2024, pp. 17\,868--17\,879.

\bibitem{du2021vision}
G.~Du, K.~Wang, S.~Lian, and K.~Zhao, ``Vision-based robotic grasping from object localization, object pose estimation to grasp estimation for parallel grippers: a review,'' \emph{Artificial Intelligence Review}, vol.~54, no.~3, pp. 1677--1734, 2021.

\bibitem{wan2020planning}
W.~Wan, K.~Harada, and F.~Kanehiro, ``Planning grasps with suction cups and parallel grippers using superimposed segmentation of object meshes,'' \emph{IEEE Transactions on Robotics}, vol. 37:1, no.~1, pp. 166--184, 2020.

\bibitem{jain2016grasp}
S.~Jain and B.~Argall, ``Grasp detection for assistive robotic manipulation,'' in \emph{2016 IEEE International Conference on Robotics and Automation (ICRA)}, 2016, pp. 2015--2021.

\bibitem{liang2019pointnetgpd}
H.~Liang, X.~Ma, S.~Li, M.~G{\"o}rner, S.~Tang, B.~Fang, F.~Sun, and J.~Zhang, ``Pointnetgpd: Detecting grasp configurations from point sets,'' in \emph{International Conference on Robotics and Automation}, 2019.

\bibitem{mohammed2020review}
M.~Q. Mohammed, K.~L. Chung, and C.~S. Chyi, ``Review of deep reinforcement learning-based object grasping: Techniques, open challenges, and recommendations,'' \emph{IEEE Access}, vol.~8, pp. 178\,450--178\,481, 2020.

\bibitem{zhang2023cherry}
Y.~Zhang, L.~Ke, A.~Deshpande, A.~Gupta, and S.~Srinivasa, ``Cherry-picking with reinforcement learning,'' \emph{arXiv preprint arXiv:2303.05508}, vol.~15, 2023.

\bibitem{tuscher2021deep}
M.~Tuscher, J.~H{\"o}rz, D.~Driess, and M.~Toussaint, ``Deep 6-dof tracking of unknown objects for reactive grasping,'' in \emph{IEEE International Conference on Robotics and Automation}.\hskip 1em plus 0.5em minus 0.4em\relax IEEE, 2021.

\bibitem{huang2022parallel}
B.~Huang, A.~Boularias, and J.~Yu, ``Parallel monte carlo tree search with batched rigid-body simulations for speeding up long-horizon episodic robot planning,'' in \emph{2022 IEEE/RSJ International Conference on Intelligent Robots and Systems (IROS)}, 2022.

\bibitem{morrison2018closing}
D.~Morrison, P.~Corke, and J.~Leitner, ``Closing the loop for robotic grasping: A real-time, generative grasp synthesis approach,'' \emph{arXiv preprint arXiv:1804.05172}, 2018.

\bibitem{marturi2019dynamic}
N.~Marturi, M.~Kopicki, A.~Rastegarpanah, V.~Rajasekaran, M.~Adjigble, R.~Stolkin, A.~Leonardis, and Y.~Bekiroglu, ``Dynamic grasp and trajectory planning for moving objects,'' \emph{Autonomous Robots}, vol.~43, no.~5, pp. 1241--1256, 2019.

\bibitem{akinola2021dynamic}
I.~Akinola, J.~Xu, S.~Song, and P.~K. Allen, ``Dynamic grasping with reachability and motion awareness,'' in \emph{IEEE/RSJ International Conference on Intelligent Robots and Systems}.\hskip 1em plus 0.5em minus 0.4em\relax IEEE, 2021.

\bibitem{wu2022survey}
J.~Wu, Z.~Jin, A.~Liu, L.~Yu, and F.~Yang, ``A survey of learning-based control of robotic visual servoing systems,'' \emph{Journal of the Franklin Institute}, vol. 359, no.~1, pp. 556--577, 2022.

\bibitem{burgess2023architecture}
B.~Burgess-Limerick, C.~Lehnert, J.~Leitner, and P.~Corke, ``An architecture for reactive mobile manipulation on-the-move,'' in \emph{2023 IEEE International Conference on Robotics and Automation (ICRA)}.\hskip 1em plus 0.5em minus 0.4em\relax IEEE, 2023, pp. 1623--1629.

\bibitem{song2020grasping}
S.~Song, A.~Zeng, J.~Lee, and T.~Funkhouser, ``Grasping in the wild: Learning 6dof closed-loop grasping from low-cost demonstrations,'' \emph{IEEE Robotics and Automation Letters}, vol. 5(3), pp. 4978--4985, 2020.

\bibitem{kalashnikov2018scalable}
D.~Kalashnikov, A.~Irpan, P.~Pastor, J.~Ibarz, A.~Herzog, E.~Jang, D.~Quillen, E.~Holly, M.~Kalakrishnan, V.~Vanhoucke \emph{et~al.}, ``Scalable deep reinforcement learning for vision-based robotic manipulation,'' in \emph{Conference on Robot Learning}, 2018.

\bibitem{r2d2}
J.~Revaud, C.~De~Souza, M.~Humenberger, and P.~Weinzaepfel, ``R2d2: Reliable and repeatable detector and descriptor,'' \emph{Advances in neural information processing systems}, vol.~32, 2019.

\bibitem{contact_grasp_net}
M.~Sundermeyer, A.~Mousavian, R.~Triebel, and D.~Fox, ``Contact-graspnet: Efficient 6-dof grasp generation in cluttered scenes,'' in \emph{2021 IEEE International Conference on Robotics and Automation (ICRA)}.\hskip 1em plus 0.5em minus 0.4em\relax IEEE, 2021, pp. 13\,438--13\,444.

\bibitem{ppo}
J.~Schulman, F.~Wolski, P.~Dhariwal, A.~Radford, and O.~Klimov, ``Proximal policy optimization algorithms,'' \emph{arXiv preprint arXiv:1707.06347}, 2017.

\bibitem{gdino}
S.~Liu, Z.~Zeng, T.~Ren, F.~Li, H.~Zhang, J.~Yang, Q.~Jiang, C.~Li, J.~Yang, H.~Su \emph{et~al.}, ``Grounding dino: Marrying dino with grounded pre-training for open-set object detection,'' in \emph{European Conference on Computer Vision}.\hskip 1em plus 0.5em minus 0.4em\relax Springer, 2024, pp. 38--55.

\bibitem{wen2023bundlesdf}
B.~Wen, J.~Tremblay, V.~Blukis, S.~Tyree, T.~M{\"u}ller, A.~Evans, D.~Fox, J.~Kautz, and S.~Birchfield, ``Bundlesdf: Neural 6-dof tracking and 3d reconstruction of unknown objects,'' in \emph{Proceedings of the IEEE/CVF Conference on Computer Vision and Pattern Recognition}, 2023, pp. 606--617.

\bibitem{sundermeyer2021contact}
M.~Sundermeyer, A.~Mousavian, R.~Triebel, and D.~Fox, ``Contact-graspnet: Efficient 6-dof grasp generation in cluttered scenes,'' 2021.

\bibitem{ekf}
K.~Fujii, ``Extended kalman filter,'' \emph{Refernce Manual}, vol.~14, p.~41, 2013.

\bibitem{2dTracker}
C.~Mayer, M.~Danelljan, G.~Bhat, M.~Paul, D.~P. Paudel, F.~Yu, and L.~Van~Gool, ``Transforming model prediction for tracking,'' in \emph{Proceedings of the IEEE/CVF conference on computer vision and pattern recognition}, 2022, pp. 8731--8740.

\bibitem{isaacgym}
V.~Makoviychuk, L.~Wawrzyniak, Y.~Guo, M.~Lu, K.~Storey, M.~Macklin, D.~Hoeller, N.~Rudin, A.~Allshire, A.~Handa \emph{et~al.}, ``Isaac gym: High performance gpu-based physics simulation for robot learning,'' \emph{arXiv preprint arXiv:2108.10470}, 2021.

\end{thebibliography}

\end{document}